\documentclass[5p,twocolumn]{elsarticle}

\usepackage{amsmath}
\usepackage[noend]{algpseudocode}
\usepackage{algorithmicx,algorithm}
\usepackage{hyperref}
\usepackage{autobreak}
\usepackage{color}
\usepackage{ulem}       
\usepackage{makecell}

\usepackage{graphicx}
\usepackage{picins}

\usepackage{bm}
\usepackage{booktabs}
\usepackage{amsfonts}
\usepackage{amsmath}
\usepackage{MnSymbol}
\usepackage{amsthm}
\usepackage{bbm}
\usepackage{color}
\usepackage{graphicx}
\usepackage{subfigure}
\usepackage{bbding}
\usepackage{url}

\usepackage{multirow}
\usepackage[switch]{lineno}
\usepackage[]{algorithm}
\usepackage{algpseudocode}

\def\+#1{\mathbb{#1}}
\def\*#1{\mathbf{#1}}
\usepackage{subfigure}  
\modulolinenumbers[5]

\journal{Journal of \LaTeX\ Templates}

\begin{document}

\begin{frontmatter}

\title{Attention-based Cross-Layer Domain Alignment for Unsupervised Domain Adaptation}

\author[zju]{Xu Ma}
\author[zju]{Junkun Yuan}
\author[ruk]{Yen-wei Chen}
\author[zju]{Ruofeng Tong}

\author[zju]{Lanfen Lin\corref{mycorrespondingauthor}}
\cortext[mycorrespondingauthor]{Corresponding author}
\ead{llf@zju.edu.cn}

\address[zju]{College of Computer Science and Technology, Zhejiang University, Hangzhou 310027, China}
\address[ruk]{College of Information Science and Engineering, Ritsumeikan University, Kusatsu 5250058, Japan}

\begin{abstract}
Unsupervised domain adaptation (UDA) aims to learn transferable knowledge from a labeled source domain and adapts a trained model to an unlabeled target domain. To bridge the gap between source and target domains, one prevailing strategy is to minimize the distribution discrepancy by aligning their semantic features extracted by deep models. The existing alignment-based methods mainly focus on reducing domain divergence in the same model layer. However, the same level of semantic information could distribute across model layers due to the domain shifts. To further boost model adaptation performance, we propose a novel method called Attention-based Cross-layer Domain Alignment (ACDA), which captures the semantic relationship between the source and target domains across model layers and calibrates each level of semantic information automatically through a dynamic attention mechanism. An elaborate attention mechanism is designed to reweight each cross-layer pair based on their semantic similarity for precise domain alignment, effectively matching each level of semantic information during model adaptation. Extensive experiments on multiple benchmark datasets consistently show that the proposed method ACDA yields state-of-the-art performance. 
\end{abstract}

\end{frontmatter}

\section{Introduction}\label{sec:int}
Deep learning has achieved remarkable progress in diverse areas of computer vision like visual recognition \cite{he2016deep} and many more. The training of deep learning models heavily relies on the independent and identical distributed (i.i.d.) assuming that training and test datasets should have the same statistical distribution. However, in real world applications, usually models trained on one dataset face test datasets which have totally different and distinct data distributions. This results in severe performance degradation because the features of the datasets are completely different from each other. Unsupervised domain adaptation (UDA) \cite{ben2010theory, ding2018graph, luo2020unsupervised, zhu2021transferable} is introduced to tackle the distribution/domain shift problem by learning transferable knowledge from a labeled source (training) domain and adapting the model to an unlabeled target (test) domain.

\begin{figure}[t]
    \centering
    \includegraphics[trim={0cm 0cm 0cm 0cm},clip,width=0.99\columnwidth]{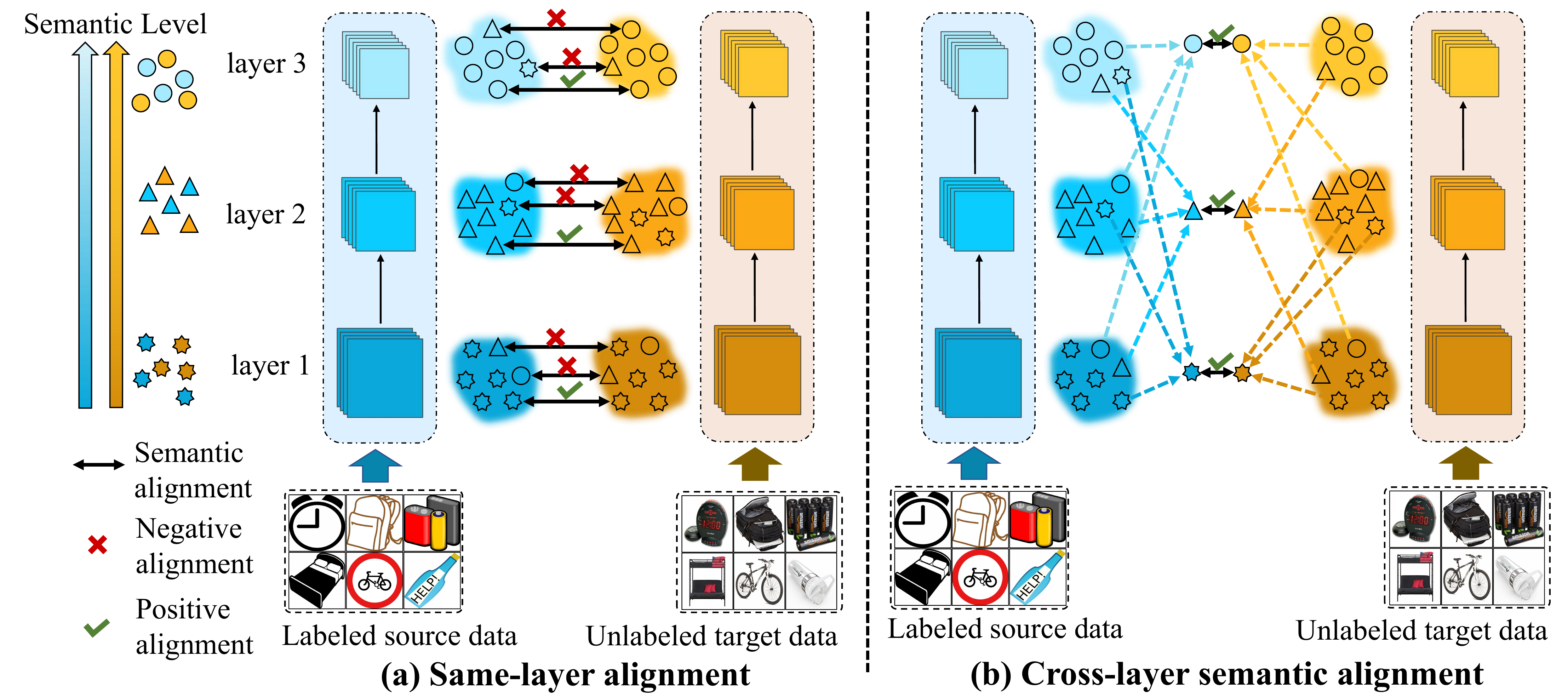}
    \caption{Comparison between (a) same-layer alignment adopted by previous methods and (b) cross-layer semantic alignment introduced by this work. Instead of directly aligning same-layer features, we match the same level of semantic information across model layers of source and target to further facilitate domain divergence minimization and improve model adaptation performance.}\label{fig:network}
\end{figure}

One prevailing strategy for UDA is to minimize the distribution discrepancy of the source and target data by aligning their semantic features extracted by deep models. However, the existing alignment-based algorithms \cite{long2015learning, long2016unsupervised, long2017deep, venkateswara2017deep} mostly reduce domain divergence by matching semantic features of the source and target data in the same layer of model, while recent studies \cite{chen2021cross, yuan2021collaborative} have shown that the same level of semantic information could distribute across model layers due to domain shifts, which we call the semantic dislocation problem. This problem makes different levels of semantic information be mismatched during the same-layer feature alignment process in the previous methods, bringing negative transfer gain to model adaptation performance. 
In comparison, we propose to match the same level of semantic information of the source and target domains across model layers for precise domain alignment, as shown in Figure \ref{fig:network}.

In order to further facilitate accurate semantic alignment and minimize domain divergence, we propose a novel method called \textit{Attention-based Cross-layer Domain Alignment (ACDA)}. This method captures and calibrates the semantic relationship between the source and target domains across the different layers of the model. Specifically, we first minimize divergence between each pair of the extracted semantic features of the source and target data. To calibrate each level of semantic information, a dynamic attention mechanism is designed to reweight the divergence minimization loss of each pair of the cross-layer on the basis of their semantic similarities. In this way, different levels of semantic information are aligned automatically, effectively minimizing the distribution discrepancy between the source and target domains and improving model adaptation capability. Extensive experiments on various standard domain adaptation benchmark datasets, i.e., Office-31, Office-Home, ImageCLEF-DA, and VisDA-2017, show that the proposed method ACDA outperforms other state-of-the-art UDA methods. 

In summary, this paper has the following contributions:

\begin{itemize}
\item We point out the semantic dislocation problem that each level of semantic information can be distributed across different layers of the model due to the domain shifts, which could bring negative transfer gain to previous methods with same-layer semantic feature alignment. 

\item In order to address the above problem, we propose a novel method called attention-based cross-layer domain alignment to match the same level of semantics by reweighting each cross-layer pair through a semantic similarity based dynamic attention mechanism. 

\item Extensive experiments show the superior performance of our method in comparison to other state-of-the-art UDA approaches on multiple standard benchmark datasets.
\end{itemize}

The rest of this paper is organized as follows. In Section \ref{sec-rel}, related works about unsupervised domain adaptation and attention mechanism are briefly introduced. In Section \ref{sec-met}, the framework and algorithm of the proposed method are described. In Section \ref{sec-exp}, the results of experiments are reported and discussed. In Section \ref{sec-con}, we conclude the investigation with a future research recommendation.

\section{Related Work}\label{sec-rel}
\subsection{Unsupervised Domain Adaptation}
Unsupervised domain adaptation (UDA) \cite{ding2018graph, luo2020unsupervised, zhu2021transferable, 2020Generative, zuo2021challenging, xu2019larger, deng2019cluster, zhang2019domain, yuan2021learning, pan2019transferrable, kang2019contrastive, hu2020unsupervised, xu2020reliable, yuan2021domain, tang2020unsupervised, ma2019gcan} aims to adapt the model trained on a labeled source domain to an unlabeled target domain when there is distinct domain divergence. A series of UDA algorithms \cite{long2017conditional, li2019cycle,yang2019cross, zhang2018collaborative, chen2020adversarial, cui2020gradually,tzeng2017adversarial} have been proposed by employing an adversarial learning strategy where the semantic features of the source and target data are aligned for reducing domain divergence. For example, a representative framework DANN \cite{ganin2016domain} utilizes the generator to extract domain-invariant semantic features that can fool the discriminator with a gradient reversal layer. Long et al. \cite{long2017conditional} extend this framework and reduce domain divergence by considering conditional probability distributions. However, the model adaptation performance of these methods rely on the carefully designed network structure and adversarial training process, which could be unstable and inefficient \cite{mescheder2018training}.

Directly minimizing domain divergence by aligning the semantic features of the source and target domains extracted by deep models is the usual way to address the domain shifts, much attention has been paid to this approach \cite{long2015learning, long2016unsupervised, long2017deep, venkateswara2017deep, shao2018feature}. 
In these approaches the discrepancy of the semantic features of the source and target domains is reduced using the distance matrix, such as maximum mean discrepancy (MMD) \cite{long2015learning} and multi-kernel MMD (MK-MMD) \cite{long2016unsupervised, long2017deep, venkateswara2017deep}. For example, Long et al. \cite{long2015learning} propose to learn a generalizable model by matching the semantic feature embeddings of different domains. However, most of these previous alignment-based methods consider the semantic relationship in the same layer of the model, while each level of semantic information can be  distributed across the layers due to the domain  shifts. Recently, Joint adaptation network (JAN) \cite{long2017deep} considers the semantic relationship in different layers, but this method can not effectively reweight the implemented cross-layer constraint, which results in an insufficient domain adaptation. In this paper, we boost model adaptation performance by exploring the relationship of cross-layer semantic information \cite{chen2021cross}. Specifically, we introduce an attention-based semantic matching method to automatically reweight the divergence minimization loss of each pair of cross-layer.


\subsection{Attention mechanism}
In recent years, attention mechanisms \cite{bahdanau2014neural}, especially self-attention \cite{vaswani2017attention}, is widely adopted in various tasks of computer vision \cite{fu2019dual, wang2019transferable, 2020Generative}. 
The self-attention mechanism learns a representation of a sequence by reweighting each position according to its corresponding similarity/importance. 
A generic process of the self-attention consists of the following steps: (1) Obtain semantic feature embeddings i.e., query, key, and value of the original features; (2) Calculate the similarity between query and key, and normalize it to obtain the weights; (3) Use the weights to synthesize the value. For example, \cite{fu2019dual} proposes to capture rich contextual dependencies for scene segmentation by obtaining attention in both spatial and channel dimensions. It first calculates the attention map of the original semantic features by performing the inner dot product of the features, then exploits the generated weights to synthesize the original semantic features. This emphasizes their important semantic information. Inspired by it, we design a dynamic attention mechanism to automatically capture the cross-layer domain-invariant semantic relationship based on the semantic similarity in both spatial and channel dimensions. The proposed attention mechanism precisely reduces domain divergence and effectively improves model adaptation towards new target domain..  

Some attention-based DA works, such as \cite{wang2019transferable} proposes to use attention for region-level and image-level context learning by exploring relationship in original images and the semantic features; \cite{li2020spatial} introduces a spatial attention pyramid network to capture context information at different scales; \cite{chen2020generative} puts forward generative attention adversarial classification network to allows a discriminator to discriminate the transferable regions; \cite{zuo2021attention} proposes attention-based multi-source DA framework by considering domain correlations and alleviating effect of the dissimilar domains. However, these methods only consider the semantic features in the final model layer, while our work matches the same level of semantic information across model layers through the elaborate dynamic attention mechanism. 

\begin{figure*}[t]
    \centering
    \includegraphics[trim={0cm 0cm 0cm 0cm},clip,width=1.99\columnwidth]{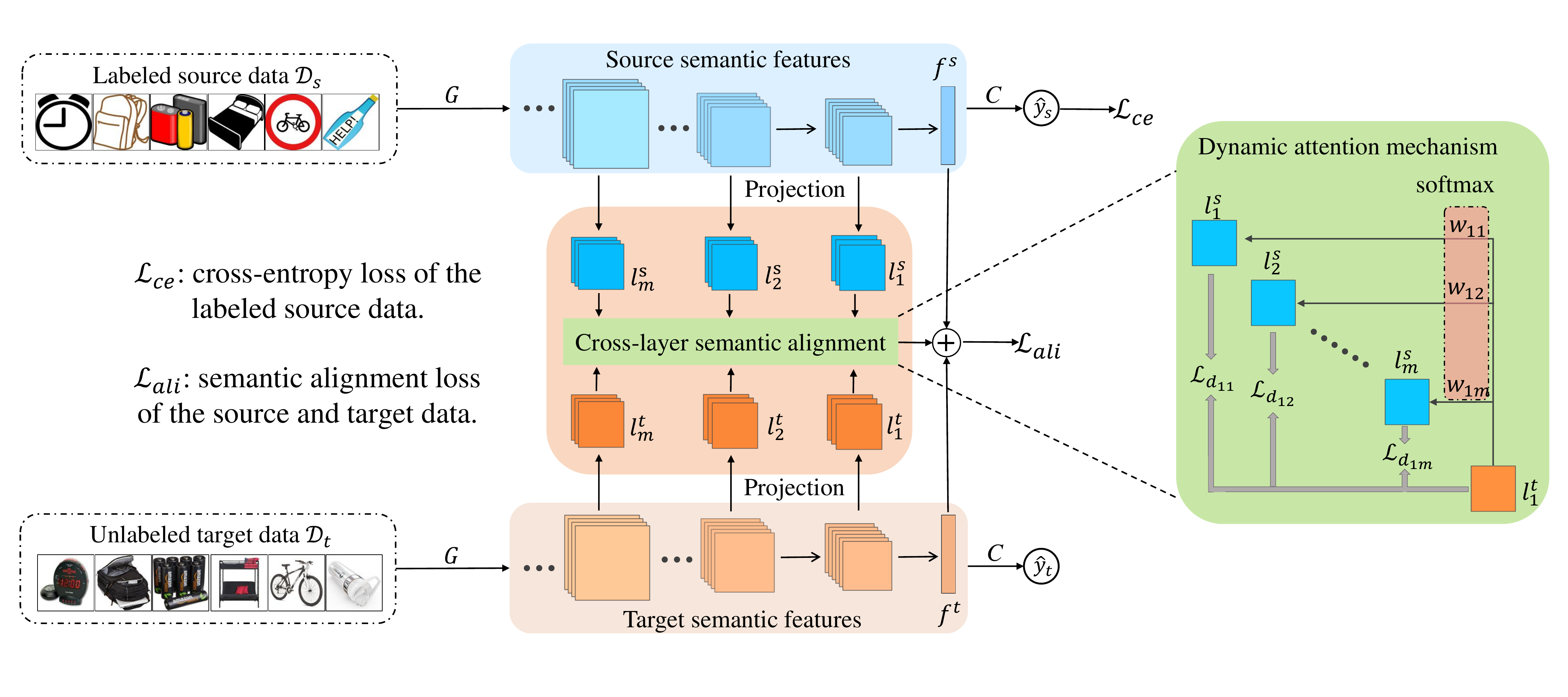}
    \caption{The proposed Attention-based Cross-layer Domain Alignment (ACDA) framework. The feature extractor $G$ extracts semantic features of the source and target data. After matching the size of feature through projection, we conduct cross-layer semantic alignment with a dynamic attention mechanism to reweight the divergence minimization loss of each cross-layer pair for precise domain alignment.}\label{fig:framework}
\end{figure*}

\section{Method}\label{sec-met}
In this section, we begin with the problem definition of unsupervised domain adaptation (UDA), and then introduce the details of the proposed method ACDA.

\subsection{Problem Definition}
Let the labeled source data be denoted as $\mathcal{D}_{s}=\left\{\left(x_{i}^{s}, y_{i}^{s}\right)\right\}_{i=1}^{n_{s}}$, where $x_{i}^s$ is the $i$-th source sample, $y_{i}^{s}$ is the corresponding class label, and $n_s$ is the source data size. The unlabeled target data is denoted as $\mathcal{D}_{t}=\left\{x_{i}^{t}\right\}_{i=1}^{n_{t}}$, where $x_{i}^t$ is the $i$-th target sample, and $n_t$ is the target data size. In UDA setting, the source and target data are sampled from different distributions $p_{s}(x, y)$ and $p_{t}(x, y)$, respectively. 
The source and target domain shares the same label space $\mathcal{C}=\{1,2, \cdots, K\}$ and we assume there are K classes in both domains.  
The goal of UDA is to train a predictive model with the labeled source data $\mathcal{D}_{s}$ and the unlabeled target data $\mathcal{D}_{{t}}$ for improving the performance of the model on the target domain.

\subsection{Cross-Layer Semantic Alignment}
Since each level of semantic information can be distributed across the layers of a model due to the domain shifts, hence we propose to match the same level semantic information of the source and target domains across the layers of the model for precise domain alignment. The framework and algorithm of the proposed ACDA method is shown in Figure \ref{fig:framework} and Algorithm \ref{alg}. The feature extractor $G$ extracts different levels of semantic features of the source and target data, and the classifier $C$ uses the high-level information of the features (i.e., top layers of a model) for object classification. 
Different from directly aligning semantic features within the same layers of the model, we learn the cross-layer semantic relationship and reweight each divergence minimization loss of each cross-layer pair according to the semantic similarity calculated using a dynamic attention mechanism. The detail is given in the following section.

\subsection{Model Pretraining}
To initialize a discriminative model, we use the labeled source data to train the model $F$, i.e., $F=C\circ G$, where $G$ and $C$ are the feature extractor and classifier, respectively.  The cross-entropy classification loss training objective for model $F$ can be defined as:
\begin{equation}\label{eq1}
    \mathcal{L}_{ce}=-\mathbb{E}_{\left(x^s, y^s\right) \sim {\mathcal{D}_s}} \sum_{k=1}^{K} \mathrm{1}_{\left[k=y^s\right]} \log \left(C_{k}\left(G(x^s)\right)\right),
\end{equation}
where $K$ is the number of classes, and $C_{k}$ is $k$-dimensional of the output of the classifier $C$. We use $\mathcal{L}_{ce}$ to pretrain the model for initializing its discrimination capability.

\begin{algorithm}[t]
    \caption{Attention-based Cross-Layer Domain Alignment}  
    \label{alg}
    \begin{algorithmic}[1]
        \Require
        The labeled source dataset $\mathcal{D}_s$ and unlabeled target dataset $\mathcal{D}_t$, the feature extractor $G$, the classifier $C$, pretraining epochs $E_{p}$ and cross-layer alignment epochs $E_{a}$, batch-size $B$.
    \Ensure  
        Trained $G$ and $C$.
    \For{$epoch=1$ to $E_{p}$} {$\quad$ // model pretraining}
        \State Sample $B$ examples from $\mathcal{D}_{s}$ and optimize $G$, $C$ by minimizing $\mathcal{L}_{ce}$ as Eq. (\ref{eq1});
    \EndFor
    \For{$epoch=1$ to $E_{a}$} $\quad$ // domain alignment
        \State Sample $B$ examples from ${\mathcal{D}}_s$ and ${\mathcal{D}}_{t}$, respectively;
        \State Feed forward the data and obtain semantic features $q^s=\{q^s_1,q^s_2,\dots,q^s_m\}$, $q^t=\{q^t_1,q^t_2,\dots,q^t_m\}$, $f^s$ and $f^t$ from encoder $G$.
        \State Obtain $l^s = \{l^s_1,l^s_2,\dots,l^s_m\}$ and  $l^t = \{l^t_1,l^t_2,\dots,l^t_m\}$ from convolution-based projections;
        \State Using $l^s$, $l^t$, $f^s$ and $f^t$ to calculate $\mathcal{L}_{ali}$ as Eq.(\ref{L-ali});
        \State Optimize $G$, $C$ and convolution-based projections by minimizing $\mathcal{L}_{all}$ as Eq. (\ref{eq12});
    \EndFor
  \end{algorithmic}
\end{algorithm}

\subsection{Cross-Layer Alignment}
After model pretraining, the initialized model can extract semantic features of source and target data. However, each level of semantic information can be distributed across layers because of the semantic dislocation problem. Therefore, we then propose to align cross-layer semantic features of the source and target domains.

\textbf{Convolution-based projection.} Since the original semantic features may have different sizes, we project the feature maps to match the size. Let the original semantic features of $m$ layers extracted by the feature extractor $G$ of the source and target are $q^s=\{q^s_1,q^s_2,...,q^s_m\}$ and $q^t=\{q^t_1,q^t_2,...,q^t_m\}$, respectively. Then we use a convolution-based projection to project all the semantic features to the same size. That is,
\begin{equation}
     l^s_i = P_{i}^{s}(q^s_i), \quad l^t_i = P_{i}^{t}(q^t_i), i=1,2,...,m
\end{equation}
where $P_{i}^{s}$ and $P_{i}^{t}$ are the convolution mapping function of the $i$-th to last layer for the source and target domains, respectively (see details in experiment section). After the projection, all the semantic features are in the same size, which allows us to implement cross-layer semantic alignment. 

\textbf{Cross-layer semantic alignment.} Since each level of semantic information could distribute across layers, we propose to align each pair of cross-layer semantic features. For simplicity, let the semantic features extracted from the target and source domain in the convolution layers (after projection) be $l^t_i$ and $l^s_j$, respectively;
and the final semantic features extracted from the target and source domains using  fully-connected layer are $f^t$ and $f^s$, respectively. 
For each cross-layer semantic feature pair $(i,j)$ of target and source domains, we minimize the domain divergence with the distance matrix:

\begin{equation}
    \begin{aligned}
        \mathrm{dist}(l^t_i,l^s_j)=\frac{1}{b_{s}^{2}} \sum_{t=1}^{b_{s}} \sum_{u=1}^{b_{s}} k\left(l_{j,t}^{s}, l_{j,u}^{s}\right) \\
        +\frac{1}{b_{t}^{2}} \sum_{t=1}^{b_{t}} \sum_{u=1}^{b_{t}} k\left(l_{i,t}^{t}, l_{i,u}^{t}\right)\\
        -\frac{2}{b_{s} b_{t}} \sum_{t=1}^{b_{s}} \sum_{u=1}^{b_{t}} k\left(l_{j,t}^{s}, l_{i,u}^{t}\right),
    \end{aligned}
\end{equation}

\begin{figure*}
    \centering
     \scalebox{0.35}[0.35]{
        \includegraphics{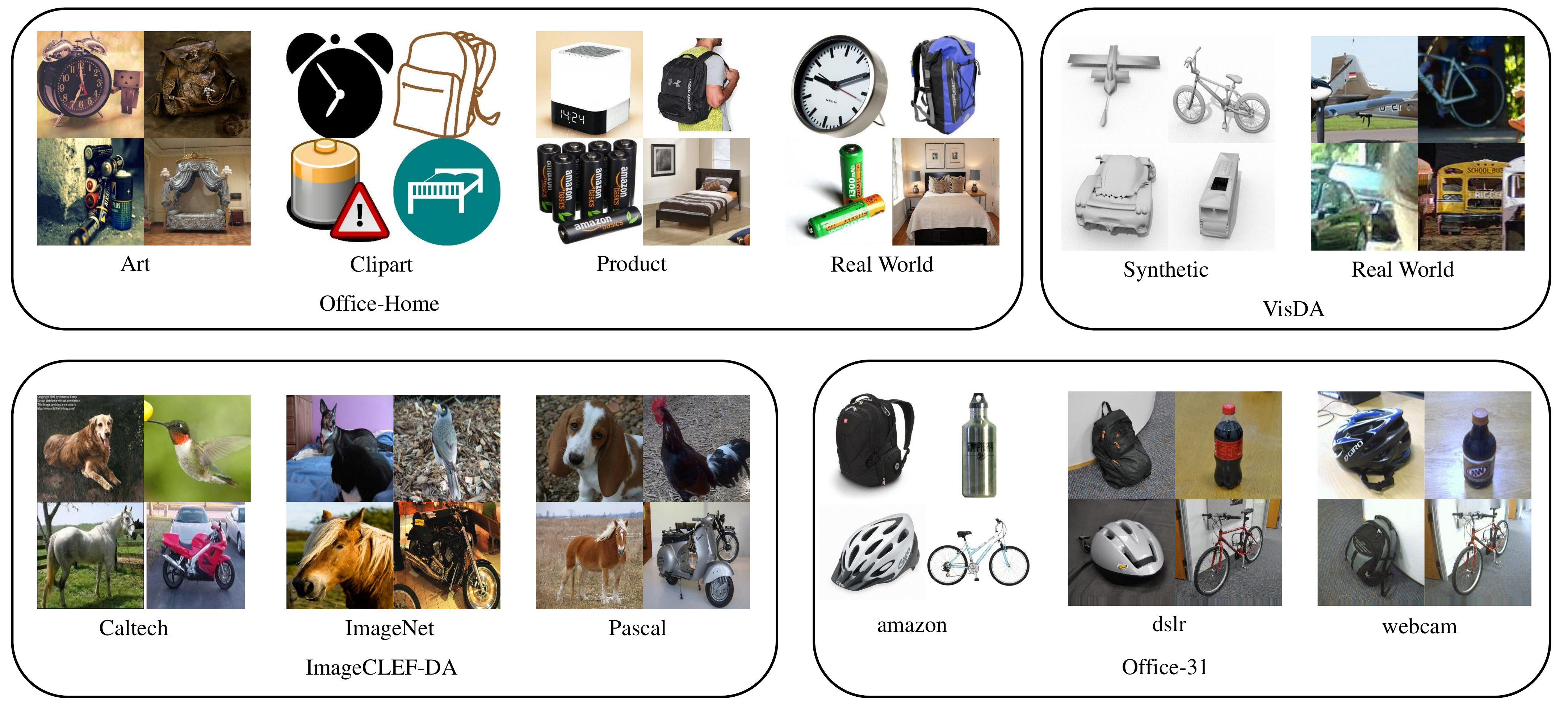}
    }
    \caption{Image examples of Office-31, Office-Home, ImageCLEF-DA, and VisDA datasets.}
    \label{fig:dataset}
\end{figure*}

where $k\left(a_1, a_2\right)=\left\langle\phi\left(a_1\right), \phi\left(a_2\right))\right\rangle$ is a kernel function, $b_s$ and $b_t$ are batch-sizes of source and target data, respectively, respectively. A characteristic kernel $k(l_j^s,l_i^t)=\left\langle\phi(l_j^s),\phi(l_i^t)\right\rangle$ is defined as a convex combination of $o$ positive semi-definite kernels $\{k_{u}\}$, i.e., $\mathcal{K}\triangleq\left\{k=\sum_{u=1}^{o}\beta_{u}k_{u}:\sum_{u=1}^{o}\beta_{u}=1,\beta_{u}>=0,\forall{u}\right\}$ \cite{long2015learning, long2017deep}. 
By minimizing the cross-layer semantic feature pairs of source and target data, we reduce the domain divergence between the source and target domains. 

\textbf{Attention allocation module.} Since the same level of semantic information could be contained in different layers, we design a dynamic attention mechanism to automatically reweight the divergence minimization loss of each cross-layer pair. Specifically, we reweight each pair of semantic features according to their semantic similarity in both spacial and channel dimensions, that is,
\begin{equation}\label{equ-att}
    \begin{aligned}
        w_{i, j}=\frac{1}{2}\frac{\exp \left[\operatorname{avg}\left(r\left(l^{t}_{i}\right) \cdot r\left(l^{s}_{j}\right)^T\right)\right]}{\sum^m_{u=1} \exp \left[\operatorname{avg}\left(r\left(l^{t}_{i}\right) \cdot r\left(l^{{s}}_{u}\right)^T\right)\right]} \\
        +\frac{1}{2}\frac{\exp \left[\operatorname{avg}\left(r\left(l^{{t}}_{i}\right)^T \cdot r\left(l^{{s}}_{j}\right)\right)\right]} {\sum^m_{u=1} \exp \left[\operatorname{avg}\left(r\left(l^{{t}}_{i}\right)^T \cdot r\left(l^{{s}}_{u}\right)\right)\right]}
    \end{aligned}
\end{equation}
where $r(\cdot)$ is a reshaping operation that maps semantic features with size $c\times{h}\times{w}$ to the size $c\times({h\times{w}})$. The \textit{avg} operator shows the global average pooling operation. We take the average of of each attention similarity matrices generating a real number value, which represents the average semantic similarity between each cross-layer pair. 
The first term and the second term of Eq. \ref{equ-att} illustrate the spatial and channel semantic relationships, respectively. 
We normalize the final similarity and obtain the weight $w_{i,j}$ for each cross-layer pair $(i,j)$, which is used to reweight the cross-layer divergence minimization loss:
\begin{equation}
    \mathcal{L}_{cross-ali}=\sum_{i} \sum_{j} w_{i j} \cdot \mathrm{dist}\left(l^{t}_{i}, l^{s}_{j}\right),
\end{equation}
where $\mathcal{L}_{cross-ali}$ is cross-layer semantic alignment loss of features. We also minimize the divergence of logits (semantic features in the final fully-connected layer from $G$), i.e., $f^s$ and $f^t$, that is,
\begin{equation}
    \mathcal{L}_{same-ali} = \mathrm{dist}(f^s,f^t).
\end{equation}
Finally, we integrate the cross-layer semantic alignment constraint of features in the convolution layers, i.e., $ \mathcal{L}_{cross-ali}$, and the same-layer semantic alignment constraint of logits, i.e., $\mathcal{L}_{same-ali}$ into a unified semantic alignment loss $\mathcal{L}_{ali}$:
\begin{equation}
    \label{L-ali}
     \mathcal{L}_{ali} = \delta \mathcal{L}_{cross-ali} + (1-\delta) \mathcal{L}_{same-ali},
\end{equation}
where $\delta$ is a trade-off hyper-parameter. Note that to further facilitate domain divergence minimization, we introduce label-conditioned cross-layer semantic alignment. Inspired by the recent work \cite{liang2020we}, we obtain pseudo labels of the unlabeled target data using k-means clustering and align the cross-layer semantic feature pairs in the same class. 

\subsection{Optimization}
The optimization of our method consists of two steps as stated in Algorithm \ref{alg}. The first step is to pretrain the model with the labeled source data by using objective loss function given in Eq. \ref{eq1}. The second step is to implement cross-layer semantic alignment by using the combination of supervised loss and alignment loss, that is,
\begin{equation}\label{eq12}
    \mathcal{L}_{all}=\mathcal{L}_{ce}+\lambda \cdot\mathcal{L}_{ali},
\end{equation}
where $\lambda$ is a hyper-parameter to keep an appropriate balance between the cross-entropy loss $\mathcal{L}_{ce}$ and the cross-layer semantic alignment loss $\mathcal{L}_{ali}$. We also analyze the sensitivity of the hyper-parameters in the experiments. 

\section{Experiments}\label{sec-exp}

\begin{table*}[t]
    \centering   
    \caption{Classification accuracy (\%) for unsupervised domain adaptation on Office-31 dataset (mean $\pm$ standard error over 3 runs).}   
    \scalebox{0.64}[0.64]{
    \renewcommand\tabcolsep{24.0pt}
        \begin{tabular}{l|cccccc|c}     
         \toprule   
            Method & A$\rightarrow$W & D$\rightarrow$W & W$\rightarrow$D & A$\rightarrow$D & D$\rightarrow$A & W$\rightarrow$A & Avg.\\
            \midrule
            ResNet-50 \cite{he2016deep} & 68.4±0.2 & 96.7±0.1 & 99.3±0.1 & 68.9±0.2 & 62.5±0.3 & 60.7±0.3 & 76.2 \\
            DAN \cite{long2015learning} & 80.5±0.4 & 97.1±0.2 & 99.6±0.1 & 78.6±0.2 & 63.6±0.3 & 62.8±0.2 & 80.4 \\
            DANN \cite{ganin2016domain} & 82.0±0.4 & 96.9±0.2 & 99.1±0.1 & 79.7±0.4 & 68.2±0.4 & 67.4±0.5 & 82.2 \\
            JAN \cite{long2017deep} & 85.4±0.3 & 97.4±0.2 & 99.8±0.2 & 84.7±0.3 & 68.6±0.3 & 70.0±0.4 & 84.3 \\
            CBST \cite{zou2018unsupervised} & 87.8±0.8 & 98.5±0.1 & \textbf{100.0±0.0} & 86.5±1.0 & 71.2±0.4 & 70.9±0.7 & 85.8 \\
            MCD \cite{saito2018maximum} & 89.6±0.2 & 98.5±0.1 & \textbf{100.0±0.0} & 91.3±0.2 & 69.6±0.1 & 70.8±0.3 & 86.6 \\
            CDAN \cite{long2017conditional} & 93.1±0.2 & 98.2±0.2 & \textbf{100.0±0.0} & 89.8±0.3 & 70.1±0.4 & 68.0±0.4 & 86.6 \\
            BSP+DANN \cite{chen2019transferability} & 93.0±0.2 & 98.0±0.2 & \textbf{100.0±0.0} & 90.0±0.4 & 71.9±0.3 & 73.0±0.3 & 87.7 \\
            GCAN \cite{ma2019gcan} & 82.7±0.1 & 97.1±0.1 & 99.8±0.1 & 76.4±0.5 & 64.9±0.1 & 62.6±0.3 & 80.6 \\
            SAFN+ENT* \cite{xu2019larger} & 90.3 & 98.7 & \textbf{100.0} & 92.1 & 73.4 & 71.2 & 87.6 \\
            rRevGrad+CAT \cite{deng2019cluster} & 94.4±0.1 & 98.0±0.2 & \textbf{100.0±0.0} & 90.8±1.8 & 72.2±0.6 & 70.2±0.1 & 87.6 \\
            SymNets \cite{zhang2019domain} &  90.8±0.1 & 98.8±0.3 & \textbf{100.0±0.0} & 93.9±0.5 & 74.6±0.6 & 72.5±0.5 & 88.4 \\
            SRDC \cite{tang2020unsupervised} & \textbf{95.7±0.2} & \textbf{99.2±0.1} & \textbf{100.0±0.0} & 95.8±0.2 & 76.7±0.3 & 77.1±0.1 & 90.8 \\
            GSDA \cite{hu2020unsupervised} & \textbf{95.7} & 99.1 & \textbf{100.0} & 94.8 & 73.5 & 74.9 & 89.7 \\
            GAACN  \cite{2020Generative} & 90.2 & 98.4 & \textbf{100.0} & 90.4 & 67.4 & 67.7 & 85.6 \\
            SCA \cite{deng2020rethinking} & 93.6±0.1 & 98.0±0.2 & \textbf{100.0±0.0} & 89.5±0.1 & 72.6±0.3 & 72.4±0.3 & 87.7 \\
            CDAN+TFLGM \cite{zhu2021transferable} & 95.3±0.3 & 99.0±0.1 & \textbf{100.0±0.0} & 94.1±0.2 & 72.5±0.2 & 71.5±0.1 & 88.7 \\
            \hline
            ACDA & 94.57±0.3 & 98.48±0.49 & 99.93±0.12 & \textbf{96.44±0.27} & \textbf{78.53±0.35} & \textbf{77.70±0.91} & \textbf{90.94} \\
            \hline
        \end{tabular}
    }
    \label{table-office31}
\end{table*}

\begin{table*}[t]
    \centering
    \caption{Classification accuracy (\%) for unsupervised domain adaptation on Office-Home dataset (mean $\pm$ standard error over 3 runs).}
    \scalebox{0.66}[0.66]{
    \renewcommand\tabcolsep{5.0pt}
        \begin{tabular}{l|cccccccccccc|c}
            \toprule
            Method & Ar$\rightarrow$Cl & Ar$\rightarrow$Pr & Ar$\rightarrow$Rw & Cl$\rightarrow$Ar & Cl$\rightarrow$Pr & Cl$\rightarrow$Rw & Pr$\rightarrow$Ar & Pr$\rightarrow$Cl & Pr$\rightarrow$Rw & Rw$\rightarrow$Ar & Rw$\rightarrow$Cl & Rw$\rightarrow$Pr & Avg.\\
	        \midrule
            ResNet-50 \cite{he2016deep} & 42.5 & 50.0 & 58.0 & 37.4 & 41.9 & 46.2 & 38.5 & 42.4 & 60.4 & 53.9 & 41.2 & 59.9 & 47.7 \\
            DAN \cite{long2015learning} & 43.6 & 57.0 & 67.9 & 45.8 & 56.5 & 60.4 & 44.0 & 43.6 & 67.7 & 63.1 & 51.5 & 74.3 & 56.3 \\
            DANN \cite{ganin2016domain} & 45.6 & 59.3 & 70.1 & 47.0 & 58.5 & 60.9 & 46.1 & 43.7 & 68.5 & 63.2 & 51.8 & 76.8  & 57.6 \\
            JAN \cite{long2017deep} & 45.9 & 61.2 & 68.9 & 50.4 & 59.7 & 61.0 & 45.8 & 43.4 & 70.3 & 63.9 & 52.4 & 76.8 & 58.3 \\
            CDAN \cite{long2017conditional} & 49.0 & 69.3 & 74.5 & 54.4 & 66.0 & 68.4 & 55.6 & 48.3 & 75.9 & 68.4 & 55.4 & 80.5 & 63.8 \\
            CBST \cite{zou2018unsupervised} & 51.4 & 74.1 & 78.9 & 56.3 & 72.2 & 73.4 & 54.4 & 41.6 & 78.8 & 66.0 & 48.3 & 81.0 & 64.7 \\
            BSP+DANN \cite{chen2019transferability} & 51.4	& 68.3 & 75.9 & 56.0 & 67.8 & 68.8 & 57.0 & 49.6 & 75.8 & 70.4 & 57.1 & 80.6 & 64.9 \\
            GAKT \cite{ding2018graph} & 34.49 & 43.63 & 55.28 & 36.14 & 52.74 & 53.16 & 31.59 & 40.55 & 61.43 & 45.64 & 44.58 & 64.92 & 47.01 \\
            SAFN* \cite{xu2019larger} & \textbf{54.4} & 73.3 & 77.9 & 65.2 & 71.5 & 73.2 & 63.6 & 52.6 & 78.2 & 72.3 & 58.0 & 82.1 & 68.5 \\
            SymNets \cite{zhang2019domain} &  47.7 & 72.9 & 78.5 & 64.2 & 71.3 & 74.2 & \textbf{64.2} & 48.8 & \textbf{79.5} & \textbf{74.5} & 52.6 & 82.7 & 67.6 \\
            GCAN \cite{ma2019gcan} & 36.43 & 47.25 & 61.08 & 37.90 & 58.25 & 57.00 & 35.77 & 42.66 & 64.47 & 50.08 & 49.12 & 72.53 & 51.05 \\
            GAACN \cite{2020Generative} & 53.1 & 71.5 & 74.6 & 59.9 & 64.6 & 67.0 & 59.2 & 53.8 & 75.1 & 70.1 & 59.3 & 80.9 & 65.8 \\
            SCA \cite{deng2020rethinking} & 46.7 & 64.6 & 71.3 & 53.1 & 65.3 & 65.2 & 54.6 & 47.2 & 71.7 & 68.2 & 56.0 & 80.2 & 62.1 \\
            DRMEA \cite{luo2020unsupervised} & 52.3±0.4 & 73.0±0.6 & 77.3±0.3 & 64.3±0.3 & 72.0±0.7 & 71.8±0.5 & 63.6±0.6 & 52.7±0.7 & 78.5±0.2 & 72.0±0.1 & 57.7±0.6 & 81.6±0.2 & 68.1±0.2 \\
            CDAN+TFLGM \cite{zhu2021transferable} & 51.4 & 72.0 & 77.2 & 61.7 & 71.9 & 72.2 & 60.0 & 51.7 & 78.8 & 72.8 & 58.9 & 82.0 & 67.6 \\
            \hline
            ACDA & 53.1±0.9 & \textbf{74.8±1.2} & \textbf{82.6±0.5} & \textbf{69.8±0.8} & \textbf{75.8±1.1} & \textbf{77.4±0.1} & 63.6±0.4 & \textbf{54.7±0.8} & 78.6±0.5 & 71.6±0.3 & \textbf{60.6±0.9} & \textbf{83.2±0.7} & \textbf{70.5} \\
            \hline
        \end{tabular}
    }
    \label{table-officehome}
\end{table*}

\begin{table*}
  \centering   
  \caption{Classification accuracy (\%) for unsupervised domain adaptation on Image-CLEF dataset (mean $\pm$ standard error over 3 runs).}   
    \scalebox{0.65}[0.65]{
    \renewcommand\tabcolsep{24.0pt}
        \begin{tabular}{l|cccccc|c}     
             \toprule   
            Method & I$\rightarrow$P & P$\rightarrow$I & I$\rightarrow$C & C$\rightarrow$I & C$\rightarrow$P & P$\rightarrow$C & Avg.\\
            \midrule
            ResNet-50 \cite{he2016deep} & 74.8±0.3 & 83.9±0.1 & 91.5±0.3 & 78.0±0.2 & 65.5±0.3 & 91.2±0.3 & 80.7 \\
            DAN \cite{long2015learning} & 74.5±0.4 & 82.2±0.2 & 92.8±0.2 & 86.3±0.4 & 69.2±0.4 & 89.8±0.4 & 82.5 \\
            DANN \cite{ganin2016domain} & 75.0±0.6 & 86.0±0.3 & \textbf{96.2±0.4} & 87.0±0.5 & 74.3±0.5 & 91.5±0.6 & 85.0 \\
            JAN \cite{long2017deep} & 76.8±0.4 & 88.0±0.2 & 94.7±0.2 & 89.5±0.3 & 74.2±0.3 & 91.7±0.3 & 85.8\\
            rRevGrad+CAT \cite{deng2019cluster} & 77.2±0.2 & \textbf{91.0±0.3} & 95.5±0.3 & \textbf{91.3±0.3} & 75.3±0.6 & 93.6±0.5 & 87.3 \\
            GCAN \cite{ma2019gcan} & 68.2±0.5 & 84.1±0.2 & 92.2±0.1 & 82.5±0.1 & 67.2±0.2 & 91.3±0.1 & 80.9 \\
            GAACN \cite{2020Generative} & 77.2 & 90.3 & 95.7 & 90.2 & 77.3 & 93.3 & 87.3 \\
            \hline
            ACDA & \textbf{77.39±0.1} & 89.32±1.0 & 95.9±0.3 & 89.41±0.6 & \textbf{78.56±0.7} & \textbf{96.51±0.2} & \textbf{87.85} \\
            \hline
        \end{tabular}
    }
    \label{table-imageclef}
\end{table*}

\begin{table*}
    \setlength{\belowcaptionskip}{2.5pt}
    \centering   
    \caption{Classification accuracy (\%) for unsupervised domain adaptation on VisDA dataset.}   
    \scalebox{0.575}[0.575]{
    \renewcommand\tabcolsep{10.0pt}
        \large
        \begin{tabular}{l|cccccccccc|c}     
            \toprule   
            Method & ResNet-50 \cite{he2016deep} & DAN \cite{long2015learning} & MCD \cite{saito2018maximum} &  CDAN \cite{long2017conditional} &  DRMEA \cite{luo2020unsupervised} & GAACN \cite{2020Generative} & CTSN \cite{zuo2021challenging} & SAFN \cite{xu2019larger} &  GSDA \cite{hu2020unsupervised} &  RWOT \cite{xu2020reliable} & ACDA \\
            \midrule
            Avg. & 52.4 & 61.1 & 71.9 & 73.7 & 79.3 & 69.8 & 75.4 & 76.1 & 81.5 & 84.0 & \textbf{86.4}\\
            \hline
        \end{tabular}
    }   
    \label{table-visda}
\end{table*}

In this section, we evaluate the performance of our ACDA method by comparing it with state-of-the-art UDA methods on four publicly available datasets: Office-31, Office-Home, ImageCLEF-DA, and VisDA. We conduct analyses to provide insights on the design each component of our model.

\subsection{Dataset}
In this section, first, we introduce the datasets considered for our experiments. These datasets are Office-31, Office-Home, ImageCLEF-DA, and VisDA. Some example images are shown in the Figure \ref{fig:dataset}.

\textbf{Office-31} is a popular dataset for domain adaptation, It has 4011 images with 31 classes, these images are collected from three different areas: 1. Amazon 'A': images downloaded from amazon.com, 2. Webcame 'W': images captured using web cameras, 3. DSLR 'D': images captured using Digital SLR cameras. These images contain different photographic settings and viewpoints. 
We evaluate the domain adaptation tasks in 6 different settings: $A\rightarrow W$, $D\rightarrow W$, $W\rightarrow D$, $A\rightarrow D$, ${D\rightarrow A}$, and ${W\rightarrow A}$, in each pair of task, the former is used as the labeled source domain and the latter is used as the unlabeled target domain.

\textbf{Office-Home} is a well organized and more challenging dataset, which contains 15,500 images with 65 categories from 4 domains. In detail, Art (Ar) denotes artistic depictions for object images, Clipart (Cl) is the picture collection of clipart, Product (Pr) is object images with a clear background which is similar to Amazon category in Office-31 dataset, and Real-World (Rw) is object images collected with a regular camera. We use all possible combinations of source and target setting. We get 12 such combinations to perform the experiments.

\textbf{ImageCLEF-DA} is a benchmark dataset for ImageCLEF 2014 domain adaptation challenge, created by selecting common categories among the following three public datasets, namely, Caltrch-256 (C), ImageNet ILSVRC 2012 (I), and Pascal VOC 2012 (P). There are 50 images in each category and 600 images in each domain. We adopt six transfer tasks of domain adaptation: I$\rightarrow$P, P$\rightarrow$I, I$\rightarrow$C, C$\rightarrow$I, C$\rightarrow$P and P$\rightarrow$C.

\textbf{VisDA} is also a challenging dataset which has images from two different domains (i.e. simulated images to real images). It contains 152,397 training images and 55,388 validation images in 12 classes. We follow the training and testing protocols of \cite{saito2017asymmetric, long2017conditional}. The training of the models has been done using labeled source data and unlabeled target data. The model then test on the target data for unsupervised domain adaptation.

\subsection{Baseline Methods}
We compare our ACDA methods with the state-of-the-art unsupervised domain adaptation methods, i.e., GAKT \cite{ding2018graph}, DRMEA \cite{luo2020unsupervised}, CDAN+TFLGM \cite{zhu2021transferable}, DAN \cite{long2015learning}, JAN \cite{long2017deep}, GAACN  \cite{2020Generative},  CTSN \cite{zuo2021challenging}, SAFN+ENT* \cite{xu2019larger}, rRevGrad+CAT \cite{deng2019cluster}, SymNets \cite{zhang2019domain}, GSDA \cite{hu2020unsupervised}, RWOT \cite{xu2020reliable}, GCAN \cite{ma2019gcan}, CDAN \cite{long2017conditional}, DANN \cite{ganin2016domain}, CBST \cite{zou2018unsupervised}, 
MCD \cite{saito2018maximum}, BSP+DANN \cite{chen2019transferability} and SCA \cite{deng2020rethinking}. We show the accuracy of the methods mentioned in their published works.

\begin{figure}
    \centering
    \subfigure[]{ \label{fig:office-31-delta} 
        \includegraphics[width=0.46\columnwidth]{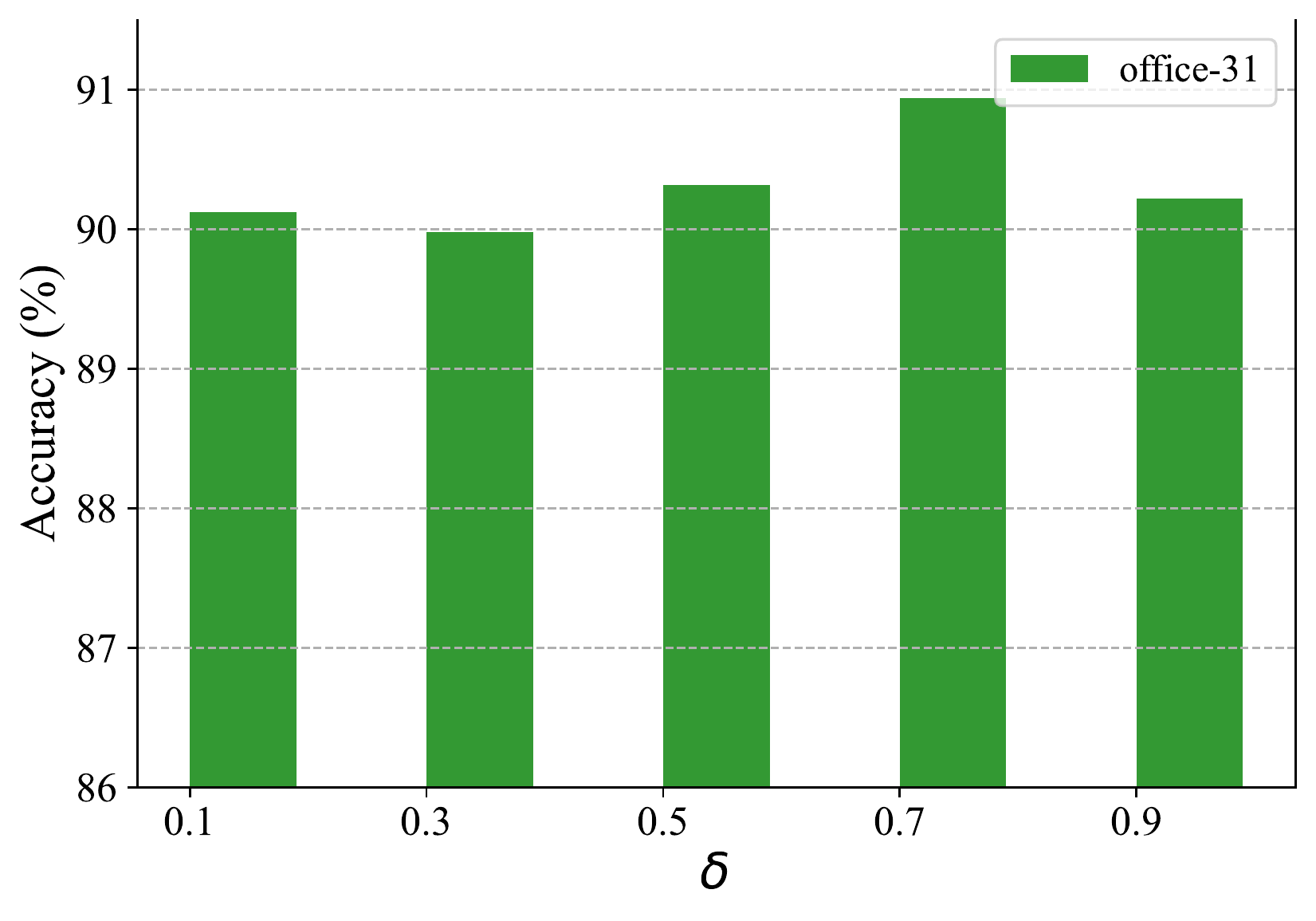} 
    }
    \subfigure[]{ \label{fig:office-31-lambda} 
        \includegraphics[width=0.46\columnwidth]{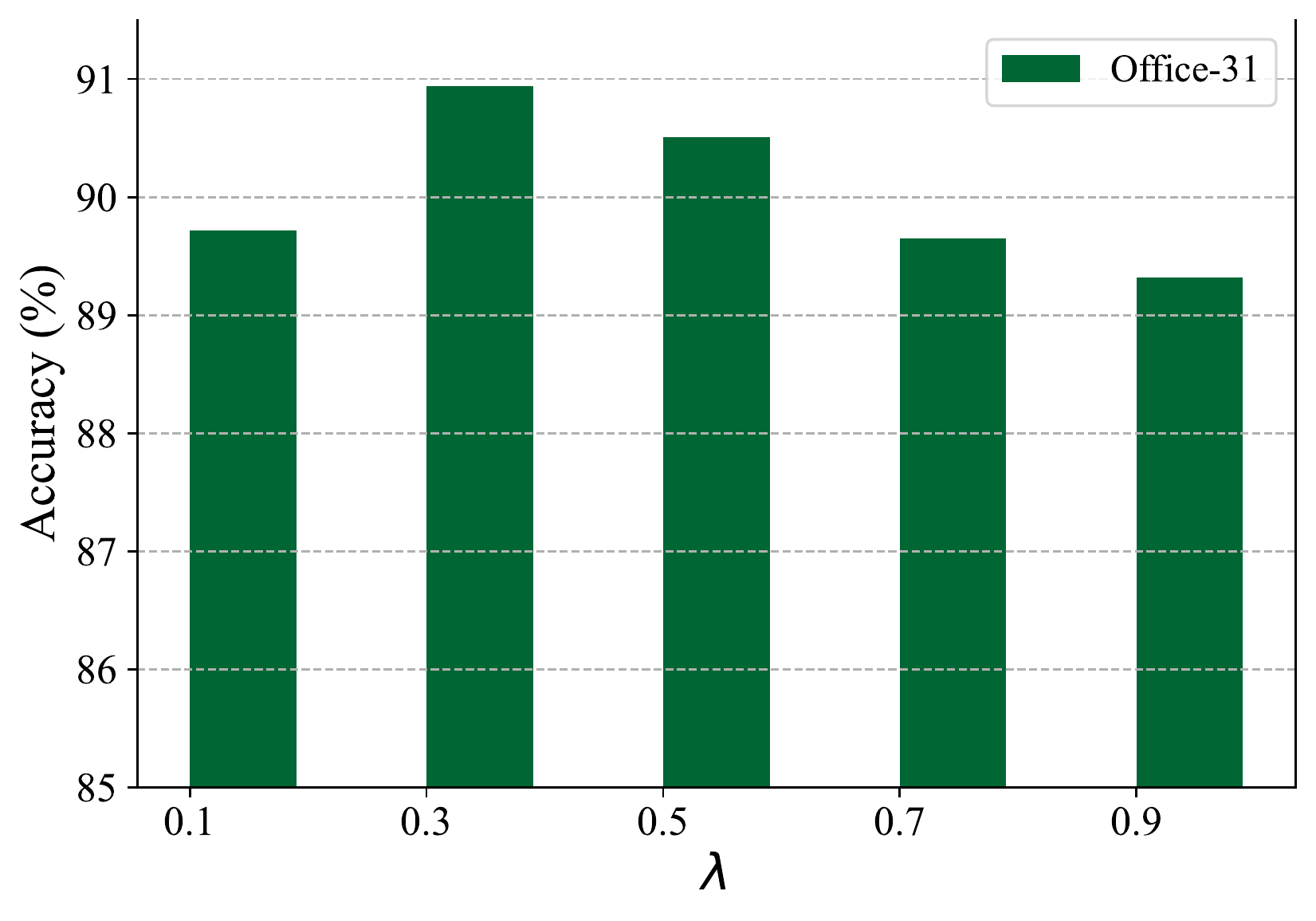} 
    }
    \caption{(a): Sensitivity analysis of hyper-parameter $\delta$ of our ACDA method on Office-31 dataset; (b): Sensitivity analysis of hyper-parameter $\lambda$ of our ACDA method on Office-31 dataset.}
    \label{fig-sensitivity}
\end{figure}

\begin{figure}
    \centering
    \subfigure[]{ \label{fig:acc} 
        \includegraphics[width=0.46\columnwidth]{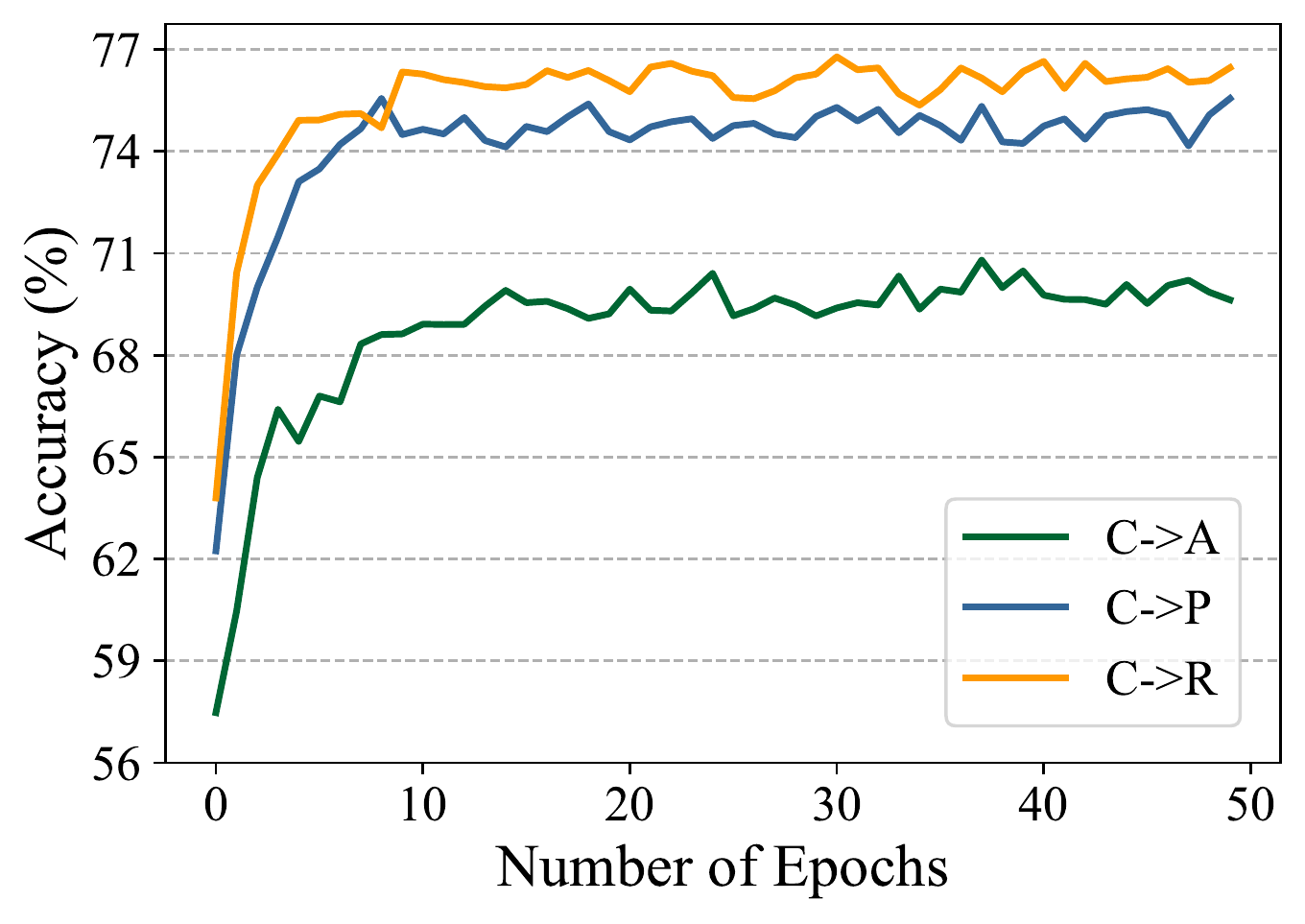} 
    } 
    \subfigure[]{ \label{fig:loss} 
        \includegraphics[width=0.46\columnwidth]{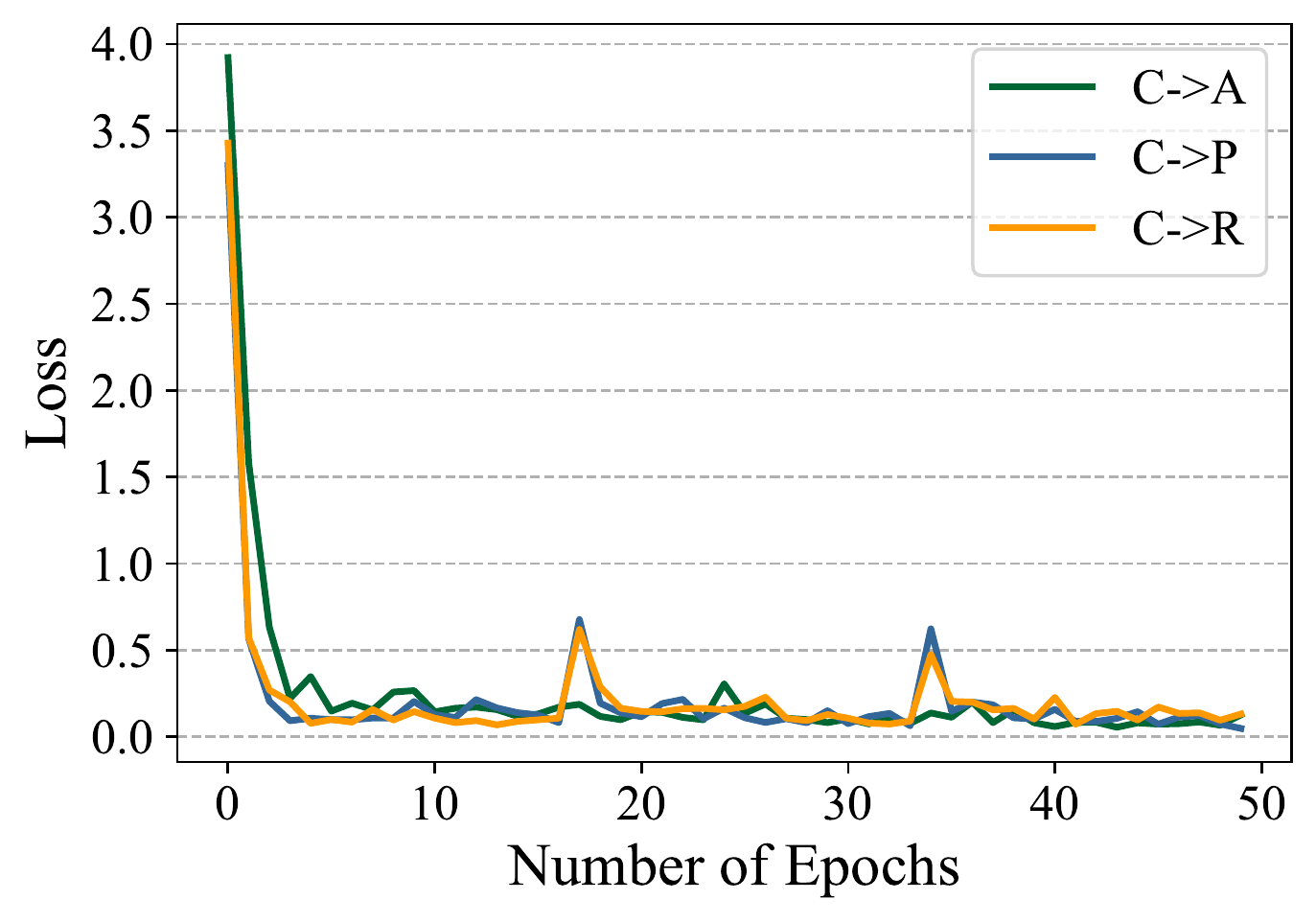} 
    }
    \caption{(a): The accuracy of our ACDA method for the transfer task C$\rightarrow$A, C$\rightarrow$P, and C$\rightarrow$R on Office-Home dataset; (b): The training loss of our ACDA method for the transfer task C$\rightarrow$A, C$\rightarrow$P, and C$\rightarrow$R on Office-Home dataset.}
    \label{fig-acc}
\end{figure}

\begin{figure*}
    \centering
    \subfigure[]{\label{fig:noadapted} 
        \includegraphics[width=0.85\columnwidth]{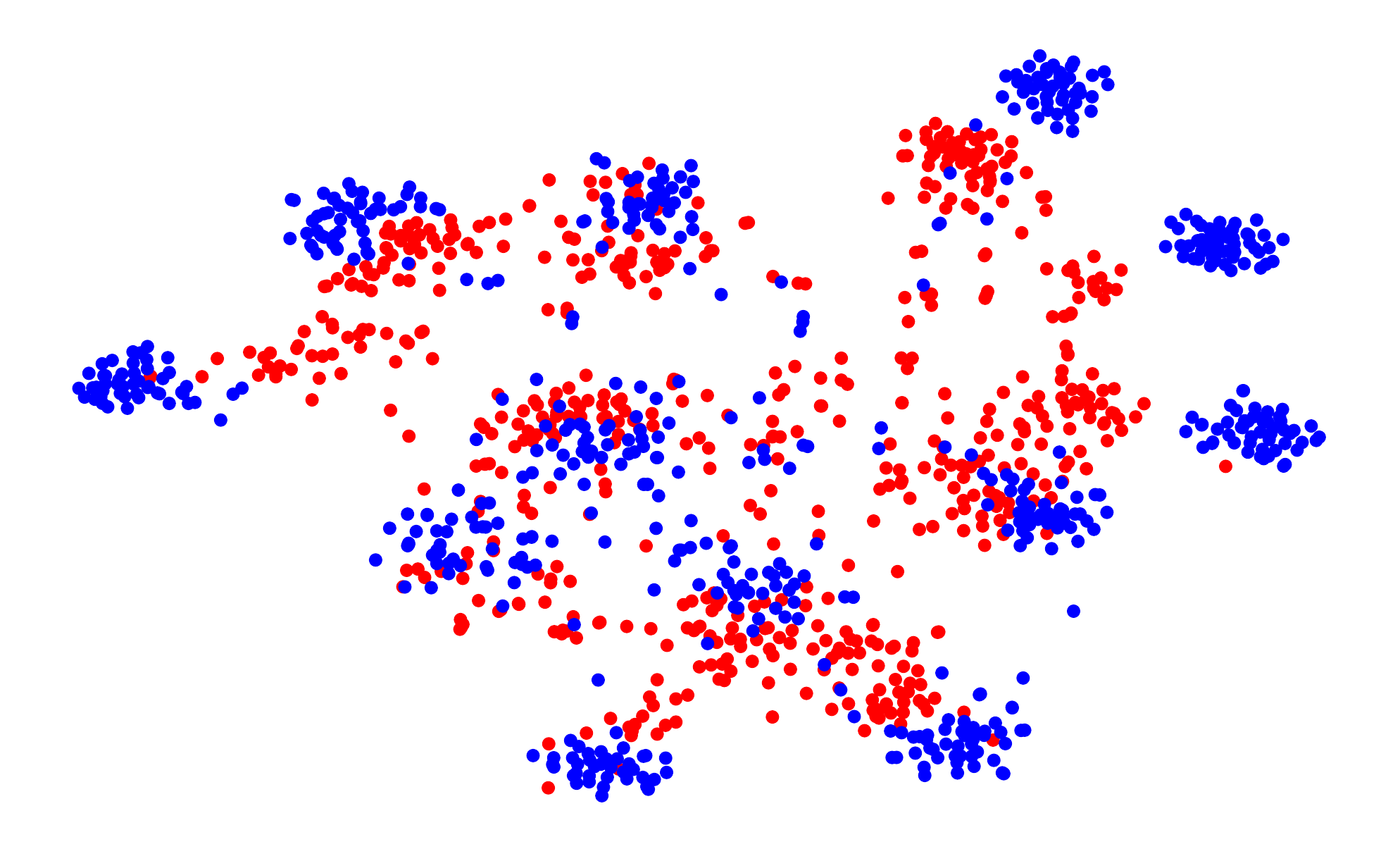} 
    } 
    \subfigure[]{\label{fig:adapted} 
        \includegraphics[width=0.85\columnwidth]{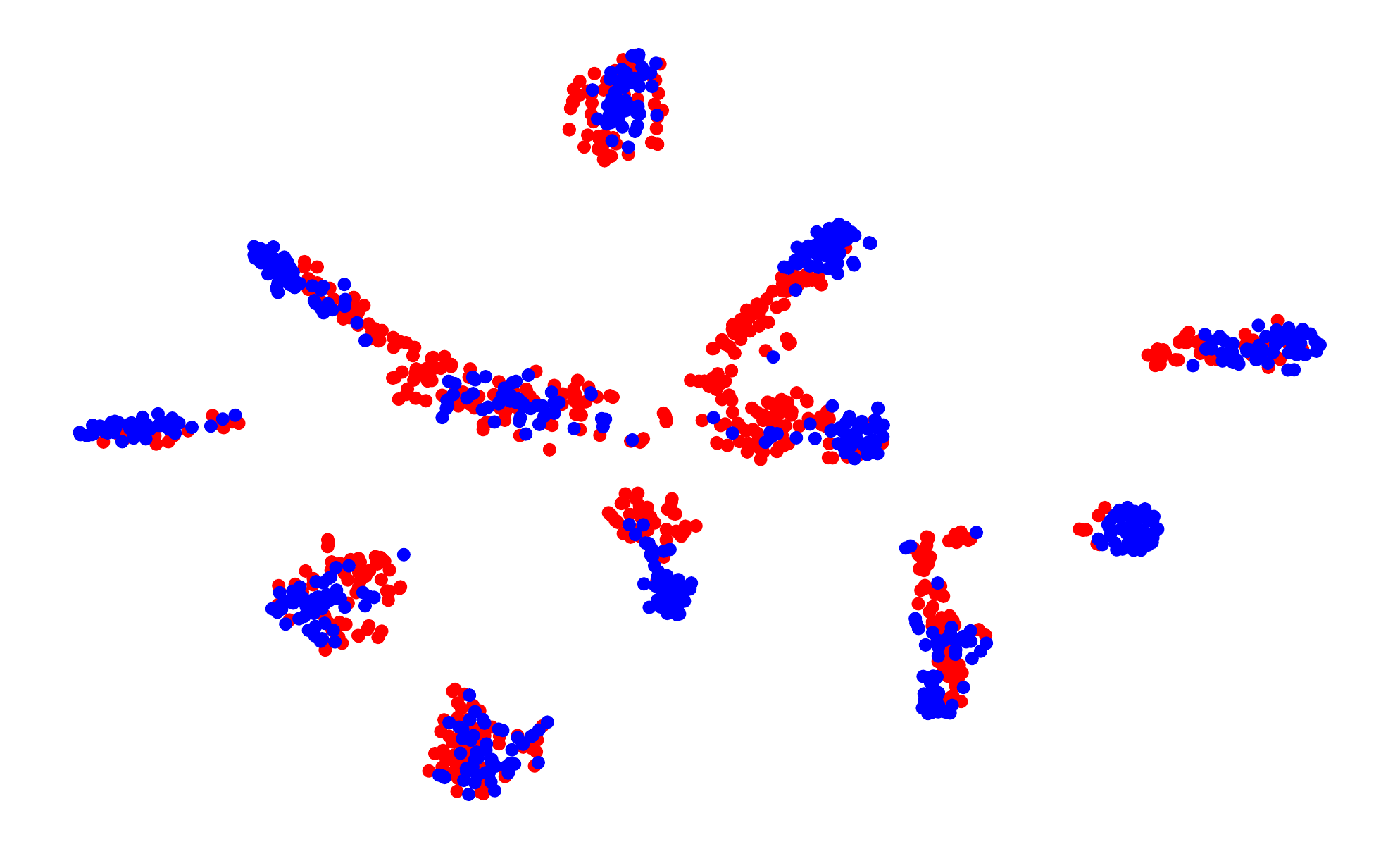} 
    }
    \caption{T-SNE visualization of the semantic feature distributions of our method for the adaptation task C$\rightarrow$P on ImageCLEF-DA dataset before adaptation (a) and after adaptation (b). The blue and red points represent samples in domain Caltech(C) and Pascal(P), respectively.}
    \label{fig:tsne}
\end{figure*}

\subsection{Implementation Details}
Following previous works \cite{long2017conditional, saito2018maximum}, we use a pretrained ResNet-50 as our backbone for the experiments on Office-31, Office-Home, and ImageCLEF-DA datasets. Whereas we use a pretrained ResNet-101 for the experiments on VisDA dataset.  We change the final fully-connected (FC) layer of the original networks with a task-specific FC layer to compose feature extractor $G$. One more FC layer is attached to it for object classification as classifier $C$. We use mini-batch stochastic gradient descent (SGD) with momentum of 0.9 to train the network. 
The semantic features used for cross-layer alignment are the features extracted by the last three blocks of the network.  
We adopt a three-layer residual convolution networks to match the size of semantic features in different layers of model. 
We set the learning rate to 0.001. We set the hyper-parameters delta and lambda to 0.7 and 0.3 in the Eq. \ref{L-ali} and Eq. \ref{eq12} to analyse the effects on performance.

\subsection{Main Results}
\textbf{Office-31}. Table \ref{table-office31} shows the average classification accuracy for the six different settings on the office-31 dataset. ACDA shows a significant improvement over all other baseline UDA methods, achieving the state-of-the-art performance. It is also worth to mention that our ACDA method achieves the best performance on half the domain adaptation setting, which indicates that our cross-layer semantic alignment strategy is significantly effective in comparison to the other alignment-based methods \cite{long2015learning, long2017deep}.

\textbf{Office-Home}. 
We report the results of the experiments on the Office-Home dataset in Table \ref{table-officehome}. 
The results show that our ACDA method outperforms other baseline methods for most of the dataset settings. Our method  improves performance for JAN by around 12.2\%, which also considers the relationship between different layer of the model. We can demonstrate it as the effectiveness of attention-based cross-layer semantic alignment strategy, which effectively calibrates each level of semantic information and facilitate precise domain adaptation. 

\textbf{ImageCLEF-DA}. The results on the ImageCLEF-DA dataset are in Table \ref{table-imageclef}. For half set of training dataset setting, our method outperforms other state-of-the-art methods and achieves the highest average classification accuracy.

\textbf{VisDA}. VisDA is a huge dataset with 152,397 and 55,388 image samples for training and validation. The experiments on VisDA dataset is reported in Table \ref{table-visda}. It is observed that our ACDA method significantly outperforms the other baseline methods on this dataset. The reason is that our ACDA method needs large data to capture and align the same level of semantic information among all the different layers of the model during training in domain adaptation setting..

\subsection{Discussions}
\textbf{Sensitivity analysis.}
In Figure \ref{fig:office-31-delta} and \ref{fig:office-31-lambda}, we report average accuracy of sensitivity analysis of the hyper-parameters $\delta$ and $\lambda$ on Office-31 dataset. We find that our ACDA method is generally robust to different hyper-parameters, indicating that exhaustive hyper-parameter fine-tuning is not very necessary for ACDA to achieve good performance. 

\begin{table}[t]
    \setlength{\belowcaptionskip}{2.5pt}
    \centering   
    \caption{Ablation experiments on Office-31 dataset for unsupervised domain adaptation.}   
    \scalebox{0.85}[0.85]{
        \small
        \begin{tabular}{p{0.85\columnwidth}|p{0.15\columnwidth}}  
            \toprule   
            Method & Avg. \\
            \midrule
            ResNet-50 \cite{he2016deep} & 76.20 \\
            ACDA w/o label-conditioned \& cross-layer alignment & 87.00 \\
            ACDA w/o cross-layer alignment & 89.27  \\
            ACDA w/o label-conditioned alignment & 88.20 \\
            ACDA w/o dynamic attention mechanism & 89.85 \\
            ACDA & \textbf{90.94} \\
            \hline
        \end{tabular}
    }   
    \label{table-ablation}
\end{table}

\begin{table}[t]
    \centering
    \caption{Comparison of different structure of convolution-based projection of ACDA method on Office-31 dataset.}
    \scalebox{0.9}[0.9]{
    \renewcommand\tabcolsep{13.0pt}
        \begin{tabular}{c|c|c|c}    
            \toprule
            Layers & Pooling & Residual block & Avg.\\
            \midrule
            \XSolidBrush & \Checkmark & \XSolidBrush & 89.78 \\
            1 & \XSolidBrush & \XSolidBrush & 90.12 \\
            3 & \XSolidBrush & \XSolidBrush & 90.29 \\
            3 & \Checkmark & \XSolidBrush & 90.48 \\
            3 & \XSolidBrush & \Checkmark & \textbf{90.94} \\
            3 & \Checkmark & \Checkmark & 90.82 \\
            \hline
        \end{tabular}
    }   
    \label{table-convolution}
\end{table}

\begin{table}[t]
    \setlength{\belowcaptionskip}{2.5pt}
    \centering
    \caption{Comparison of different number of layers for alignment on Office-31 dataset.}
    \scalebox{0.94}[0.94]{
        \renewcommand\tabcolsep{15.0pt}
        \begin{tabular}{c|c}
            \toprule
            Number of layers for alignment & Avg. \\
            \midrule
            2 & 89.87 \\
            3 & 90.94 \\
            4 & 91.02 \\
            all & \textbf{91.09} \\
            \hline
        \end{tabular}
    }   
    \label{table-cross-num}
\end{table}

\textbf{Training curves of accuracy and loss.}
Figure \ref{fig:acc} and Figure \ref{fig:loss} show the training accuracy curve and training loss curve of the three training setting on Office-Home dataset. 
It shows that our ACDA algorithm is more stable and it converges easily for different dataset settings. It indicates that the cross-layer semantic alignment strategy is very general and more robust. 

\textbf{Semantic feature distributions.}
To further analyze the effectiveness of our adaptation strategy, we visualize the learned semantic feature distributions for the C$\rightarrow$P setting on ImageCLEF-DA dataset using t-SNE \cite{van2008visualizing} in Figure \ref{fig:tsne}. It is observed that our ACDA method learns more domain-invariant semantic feature representations by aligning the source and target data in the semantic feature representation space via cross-layer semantic alignment. It proves that our method ACDA consistently achieves excellent performance for the UDA task on different datasets.

\textbf{Ablation studies.}
We also implement ablation studies to investigate the effectiveness of each part of our ACDA method. The results are reported in Table \ref{table-ablation}. 
Here, we see that the introduced label-conditioned alignment method is helpful to achieve better performance. We find that it is necessary to explore the relationship between cross-layers of source and target domains. Besides, the elaborate dynamic attention mechanism is effective for achieving state-of-the-art model adaptation performance. 

\textbf{Structure of Projection Network}
We adopt a three-layer residual convolution networks as the learnable projection network, we compare different structure of the projection networks in Table \ref{table-convolution}. It shows that the different structure of the projection networks have minor impact for the ACDA method to achieve the excellent model adaptation performance, and the adopted three-layer residual convolution networks achieves better results. 

\textbf{Alignment Layers}
We compare different number of layers for alignment in Table \ref{table-cross-num}. It shows that we may have minor adaptation performance gain by adopting more layers for alignment, but it may increase the calculation cost. 

\section{Conclusion}\label{sec-con}
In this paper, we first point out that the same level of semantic information can be distributed across the different layers of the model, which can cause of negative transfer gain in previous UDA methods with same-layer alignment. We propose a novel attention-based cross-layer domain alignment method to address this problem by reweighting each cross-layer pair according to the semantic similarity for precise domain alignment. Extensive experiments show the superior performance of our method in comparison to the other state-of-the-art UDA methods. In future, we  will extend our framework to other adaptation tasks of computer vision, like semantic segmentation and object detection.

\section*{CRediT authorship contribution statement}

\textbf{Xu Ma:} Writing - original draft, Software, Conceptualization, Methodolog, Validation. \textbf{Junkun Yuan:} Software, Writing - original draft. \textbf{Yen-Wei Chen:} Writing - review $\&$ editing. \textbf{Ruofeng Tong:} Writing - review $\&$ editing. \textbf{Lanfen Lin:} Supervision, Writing - review $\&$ editing.

\section*{Declaration of Competing Interest}
The authors declare that they have no known competing financial interests or personal relationships that could have appeared to influence the work reported in this paper.

\section*{Acknowledgments}
This work is supported by Major Scientific Project of ZhejiangLab under No. 2020ND8AD01.

\bibliography{mybibfile}

\begin{thebibliography}{10}
\expandafter\ifx\csname url\endcsname\relax
  \def\url#1{\texttt{#1}}\fi
\expandafter\ifx\csname urlprefix\endcsname\relax\def\urlprefix{URL }\fi
\expandafter\ifx\csname href\endcsname\relax
  \def\href#1#2{#2} \def\path#1{#1}\fi

\bibitem{he2016deep}
K.~He, X.~Zhang, S.~Ren, J.~Sun, Deep residual learning for image recognition,
  in: Proceedings of the IEEE conference on computer vision and pattern
  recognition, 2016, pp. 770--778.

\bibitem{ben2010theory}
S.~Ben-David, J.~Blitzer, K.~Crammer, A.~Kulesza, F.~Pereira, J.~W. Vaughan, A
  theory of learning from different domains, Machine learning 79~(1-2) (2010)
  151--175.

\bibitem{ding2018graph}
Z.~Ding, S.~Li, M.~Shao, Y.~Fu, Graph adaptive knowledge transfer for
  unsupervised domain adaptation, in: Proceedings of the European Conference on
  Computer Vision (ECCV), 2018, pp. 37--52.

\bibitem{luo2020unsupervised}
Y.-W. Luo, C.-X. Ren, P.~Ge, K.-K. Huang, Y.-F. Yu, Unsupervised domain
  adaptation via discriminative manifold embedding and alignment, in:
  Proceedings of the AAAI Conference on Artificial Intelligence, Vol.~34, 2020,
  pp. 5029--5036.

\bibitem{zhu2021transferable}
R.~Zhu, X.~Jiang, J.~Lu, S.~Li, Transferable feature learning on graphs across
  visual domains, in: 2021 IEEE International Conference on Multimedia and Expo
  (ICME), IEEE, 2021, pp. 1--6.

\bibitem{long2015learning}
M.~Long, Y.~Cao, J.~Wang, M.~Jordan, Learning transferable features with deep
  adaptation networks, in: International conference on machine learning, PMLR,
  2015, pp. 97--105.

\bibitem{long2016unsupervised}
M.~Long, H.~Zhu, J.~Wang, M.~I. Jordan, Unsupervised domain adaptation with
  residual transfer networks, arXiv preprint arXiv:1602.04433.

\bibitem{long2017deep}
M.~Long, H.~Zhu, J.~Wang, M.~I. Jordan, Deep transfer learning with joint
  adaptation networks, in: International conference on machine learning, PMLR,
  2017, pp. 2208--2217.

\bibitem{venkateswara2017deep}
H.~Venkateswara, J.~Eusebio, S.~Chakraborty, S.~Panchanathan, Deep hashing
  network for unsupervised domain adaptation, in: Proceedings of the IEEE
  conference on computer vision and pattern recognition, 2017, pp. 5018--5027.

\bibitem{chen2021cross}
D.~Chen, J.-P. Mei, Y.~Zhang, C.~Wang, Z.~Wang, Y.~Feng, C.~Chen, Cross-layer
  distillation with semantic calibration, in: Proceedings of the AAAI
  Conference on Artificial Intelligence, Vol.~35, 2021, pp. 7028--7036.

\bibitem{yuan2021collaborative}
J.~Yuan, X.~Ma, D.~Chen, K.~Kuang, F.~Wu, L.~Lin, Collaborative semantic
  aggregation and calibration for separated domain generalization, arXiv
  e-prints (2021) arXiv--2110.

\bibitem{2020Generative}
W.~Chen, H.~Hu, Generative attention adversarial classification network for
  unsupervised domain adaptation, Pattern Recognition 107 (2020) 107440.

\bibitem{zuo2021challenging}
L.~Zuo, M.~Jing, J.~Li, L.~Zhu, K.~Lu, Y.~Yang, Challenging tough samples in
  unsupervised domain adaptation, Pattern Recognition 110 (2021) 107540.

\bibitem{xu2019larger}
R.~Xu, G.~Li, J.~Yang, L.~Lin, Larger norm more transferable: An adaptive
  feature norm approach for unsupervised domain adaptation, in: Proceedings of
  the IEEE/CVF International Conference on Computer Vision, 2019, pp.
  1426--1435.

\bibitem{deng2019cluster}
Z.~Deng, Y.~Luo, J.~Zhu, Cluster alignment with a teacher for unsupervised
  domain adaptation, in: Proceedings of the IEEE/CVF International Conference
  on Computer Vision, 2019, pp. 9944--9953.

\bibitem{zhang2019domain}
Y.~Zhang, H.~Tang, K.~Jia, M.~Tan, Domain-symmetric networks for adversarial
  domain adaptation, in: Proceedings of the IEEE/CVF Conference on Computer
  Vision and Pattern Recognition, 2019, pp. 5031--5040.

\bibitem{yuan2021learning}
J.~Yuan, X.~Ma, K.~Kuang, R.~Xiong, M.~Gong, L.~Lin, Learning domain-invariant
  relationship with instrumental variable for domain generalization, arXiv
  preprint arXiv:2110.01438.

\bibitem{pan2019transferrable}
Y.~Pan, T.~Yao, Y.~Li, Y.~Wang, C.-W. Ngo, T.~Mei, Transferrable prototypical
  networks for unsupervised domain adaptation, in: Proceedings of the IEEE/CVF
  Conference on Computer Vision and Pattern Recognition, 2019, pp. 2239--2247.

\bibitem{kang2019contrastive}
G.~Kang, L.~Jiang, Y.~Yang, A.~G. Hauptmann, Contrastive adaptation network for
  unsupervised domain adaptation, in: Proceedings of the IEEE/CVF Conference on
  Computer Vision and Pattern Recognition, 2019, pp. 4893--4902.

\bibitem{hu2020unsupervised}
L.~Hu, M.~Kan, S.~Shan, X.~Chen, Unsupervised domain adaptation with
  hierarchical gradient synchronization, in: Proceedings of the IEEE/CVF
  Conference on Computer Vision and Pattern Recognition, 2020, pp. 4043--4052.

\bibitem{xu2020reliable}
R.~Xu, P.~Liu, L.~Wang, C.~Chen, J.~Wang, Reliable weighted optimal transport
  for unsupervised domain adaptation, in: Proceedings of the IEEE/CVF
  Conference on Computer Vision and Pattern Recognition, 2020, pp. 4394--4403.

\bibitem{yuan2021domain}
J.~Yuan, X.~Ma, D.~Chen, K.~Kuang, F.~Wu, L.~Lin, Domain-specific bias
  filtering for single labeled domain generalization, arXiv preprint
  arXiv:2110.00726.

\bibitem{tang2020unsupervised}
H.~Tang, K.~Chen, K.~Jia, Unsupervised domain adaptation via structurally
  regularized deep clustering, in: Proceedings of the IEEE/CVF conference on
  computer vision and pattern recognition, 2020, pp. 8725--8735.

\bibitem{ma2019gcan}
X.~Ma, T.~Zhang, C.~Xu, Gcan: Graph convolutional adversarial network for
  unsupervised domain adaptation, in: Proceedings of the IEEE/CVF Conference on
  Computer Vision and Pattern Recognition, 2019, pp. 8266--8276.

\bibitem{long2017conditional}
M.~Long, Z.~Cao, J.~Wang, M.~I. Jordan, Conditional adversarial domain
  adaptation, arXiv preprint arXiv:1705.10667.

\bibitem{li2019cycle}
J.~Li, E.~Chen, Z.~Ding, L.~Zhu, K.~Lu, Z.~Huang, Cycle-consistent conditional
  adversarial transfer networks, in: Proceedings of the 27th ACM International
  Conference on Multimedia, 2019, pp. 747--755.

\bibitem{yang2019cross}
B.~Yang, P.~C. Yuen, Cross-domain visual representations via unsupervised graph
  alignment, in: Proceedings of the AAAI Conference on Artificial Intelligence,
  Vol.~33, 2019, pp. 5613--5620.

\bibitem{zhang2018collaborative}
W.~Zhang, W.~Ouyang, W.~Li, D.~Xu, Collaborative and adversarial network for
  unsupervised domain adaptation, in: Proceedings of the IEEE conference on
  computer vision and pattern recognition, 2018, pp. 3801--3809.

\bibitem{chen2020adversarial}
M.~Chen, S.~Zhao, H.~Liu, D.~Cai, Adversarial-learned loss for domain
  adaptation, in: Proceedings of the AAAI Conference on Artificial
  Intelligence, 2020, pp. 3521--3528.

\bibitem{cui2020gradually}
S.~Cui, S.~Wang, J.~Zhuo, C.~Su, Q.~Huang, Q.~Tian, Gradually vanishing bridge
  for adversarial domain adaptation, in: Proceedings of the IEEE/CVF Conference
  on Computer Vision and Pattern Recognition, 2020, pp. 12455--12464.

\bibitem{tzeng2017adversarial}
E.~Tzeng, J.~Hoffman, K.~Saenko, T.~Darrell, Adversarial discriminative domain
  adaptation, in: Proceedings of the IEEE conference on computer vision and
  pattern recognition, 2017, pp. 7167--7176.

\bibitem{ganin2016domain}
Y.~Ganin, E.~Ustinova, H.~Ajakan, P.~Germain, H.~Larochelle, F.~Laviolette,
  M.~Marchand, V.~Lempitsky, Domain-adversarial training of neural networks,
  The journal of machine learning research 17~(1) (2016) 2096--2030.

\bibitem{mescheder2018training}
L.~Mescheder, A.~Geiger, S.~Nowozin, Which training methods for gans do
  actually converge?, in: International conference on machine learning, PMLR,
  2018, pp. 3481--3490.

\bibitem{shao2018feature}
R.~Shao, X.~Lan, P.~C. Yuen, Feature constrained by pixel: Hierarchical
  adversarial deep domain adaptation, in: Proceedings of the 26th ACM
  international conference on Multimedia, 2018, pp. 220--228.

\bibitem{bahdanau2014neural}
D.~Bahdanau, K.~Cho, Y.~Bengio, Neural machine translation by jointly learning
  to align and translate, arXiv preprint arXiv:1409.0473.

\bibitem{vaswani2017attention}
A.~Vaswani, N.~Shazeer, N.~Parmar, J.~Uszkoreit, L.~Jones, A.~N. Gomez,
  {\L}.~Kaiser, I.~Polosukhin, Attention is all you need, in: Advances in
  neural information processing systems, 2017, pp. 5998--6008.

\bibitem{fu2019dual}
J.~Fu, J.~Liu, H.~Tian, Y.~Li, Y.~Bao, Z.~Fang, H.~Lu, Dual attention network
  for scene segmentation, in: Proceedings of the IEEE/CVF Conference on
  Computer Vision and Pattern Recognition, 2019, pp. 3146--3154.

\bibitem{wang2019transferable}
X.~Wang, L.~Li, W.~Ye, M.~Long, J.~Wang, Transferable attention for domain
  adaptation, in: Proceedings of the AAAI Conference on Artificial
  Intelligence, 2019, pp. 5345--5352.

\bibitem{li2020spatial}
C.~Li, D.~Du, L.~Zhang, L.~Wen, T.~Luo, Y.~Wu, P.~Zhu, Spatial attention
  pyramid network for unsupervised domain adaptation, in: European Conference
  on Computer Vision, Springer, 2020, pp. 481--497.

\bibitem{chen2020generative}
W.~Chen, H.~Hu, Generative attention adversarial classification network for
  unsupervised domain adaptation, Pattern Recognition 107 (2020) 107440.

\bibitem{zuo2021attention}
Y.~Zuo, H.~Yao, C.~Xu, Attention-based multi-source domain adaptation, IEEE
  Transactions on Image Processing 30 (2021) 3793--3803.

\bibitem{liang2020we}
J.~Liang, D.~Hu, J.~Feng, Do we really need to access the source data? source
  hypothesis transfer for unsupervised domain adaptation, in: International
  Conference on Machine Learning, PMLR, 2020, pp. 6028--6039.

\bibitem{zou2018unsupervised}
Y.~Zou, Z.~Yu, B.~Kumar, J.~Wang, Unsupervised domain adaptation for semantic
  segmentation via class-balanced self-training, in: Proceedings of the
  European conference on computer vision (ECCV), 2018, pp. 289--305.

\bibitem{saito2018maximum}
K.~Saito, K.~Watanabe, Y.~Ushiku, T.~Harada, Maximum classifier discrepancy for
  unsupervised domain adaptation, in: Proceedings of the IEEE conference on
  computer vision and pattern recognition, 2018, pp. 3723--3732.

\bibitem{chen2019transferability}
X.~Chen, S.~Wang, M.~Long, J.~Wang, Transferability vs. discriminability: Batch
  spectral penalization for adversarial domain adaptation, in: International
  conference on machine learning, PMLR, 2019, pp. 1081--1090.

\bibitem{deng2020rethinking}
W.~Deng, L.~Zheng, Y.~Sun, J.~Jiao, Rethinking triplet loss for domain
  adaptation, IEEE Transactions on Circuits and Systems for Video Technology
  31~(1) (2020) 29--37.

\bibitem{saito2017asymmetric}
K.~Saito, Y.~Ushiku, T.~Harada, Asymmetric tri-training for unsupervised domain
  adaptation, in: International Conference on Machine Learning, PMLR, 2017, pp.
  2988--2997.

\bibitem{van2008visualizing}
L.~Van~der Maaten, G.~Hinton, Visualizing data using t-sne., Journal of machine
  learning research 9~(11).

\end{thebibliography}

    \par\noindent 
    \parbox[t]{\linewidth}{
    \noindent\parpic{\includegraphics[height=1.5in,width=1in,clip,keepaspectratio]{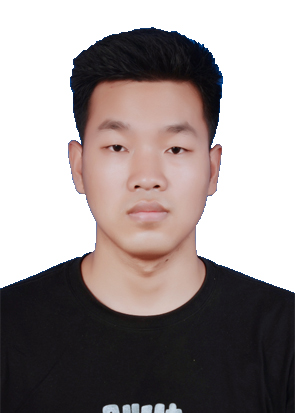}}
    \noindent \textbf{Xu Ma} received his B.S. degree from the College of Computer Science and Technology at Harbin Institute of Technology in 2020. He is currently pursuing Master degree at the College of Computer Science and Technology at Zhejiang University since 2020. His research interests include domain adaptation and domain generalization.}
    \vspace{2\baselineskip}
    
    \par\noindent 
    \parbox[t]{\linewidth}{
    \noindent\parpic{\includegraphics[height=1.5in,width=1in,clip,keepaspectratio]{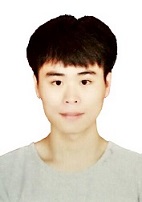}}
    \noindent \textbf{Junkun Yuan} received his B.S. degree from the College of Information Engineering at Zhejiang University of Technology in 2019.  He is currently pursuing the Ph.D. degree with the College of Computer Science and Technology at Zhejiang University since 2019. }
    \vspace{2\baselineskip}
    
    \par\noindent 
    \parbox[t]{\linewidth}{
    \noindent\parpic{\includegraphics[height=1.5in,width=1in,clip,keepaspectratio]{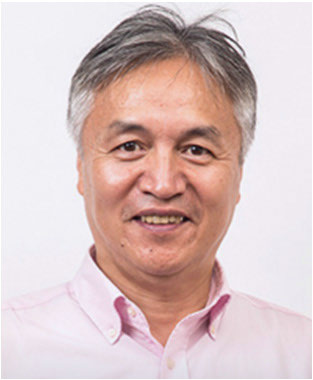}}
    \noindent \textbf{Yen-Wei Chen} (Member, IEEE) received the B.E. degree from Kobe University, Kobe, Japan, in 1985, and the M.E. and D.E. degrees from Osaka University, Osaka, Japan, in 1987 and 1990, respectively. From 1991 to 1994, he was a Research Fellow with the Institute for Laser Technology, Osaka. From October 1994 to March 2004, he was an Associate Professor and a Professor with the Department of Electrical and Electronic Engineering, University of the Ryukyus, Okinawa, Japan. He is currently a Professor with the College of Information Science and Engineering, Ritsumeikan University, Kyoto, Japan. He is also a Visiting Professor with the College of Computer Science and Technology, Zhejiang University, China, and the Research Center for Healthcare Data Science, Zhejiang Laboratory, China. His research interests include pattern recognition, image processing, and machine learning. He has published more than 200 research articles in these fields. He is an Associate Editor of the International Journal of Image and Graphics (IJIG) and an Associate Editor of the International Journal of Knowledge-Based and Intelligent Engineering Systems. }
    \vspace{1\baselineskip}
    
    \par\noindent 
    \parbox[t]{\linewidth}{
    \noindent\parpic{\includegraphics[height=1.5in,width=1in,clip,keepaspectratio]{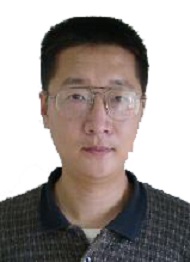}}
    \noindent \textbf{Ruofeng Tong} received his B.S. degree in mathematics from Fudan University and his Ph.D. degree in applied mathematics in 1996 from Zhejiang University. Currently, he is a professor in the College of Computer Science and Engineering, Zhejiang University, China. His research interests include CAD{\&}CG, medical image reconstruction, and virtual reality.}
    \vspace{1\baselineskip}

    \par\noindent 
    \parbox[t]{\linewidth}{
    \noindent\parpic{\includegraphics[height=1.5in,width=1in,clip,keepaspectratio]{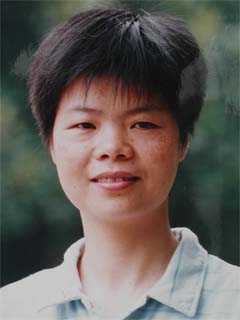}}
    \noindent \textbf{Lanfen Lin} (Member, IEEE) received Ph.D. degrees in Aircraft Manufacture Engineering from Northwestern Polytechnical University in 1995. She held a postdoctoral position with the College of Computer Science and Technology, Zhejiang University, China, from January 1996 to December 1997. Now she is a Full Professor and the Vice Director of the Artificial Intelligence Institute in Zhejiang University. Her research interests include medical image processing, big data analysis, data mining, and so on.}

\end{document}